\documentclass{ws-ijprai}
\usepackage{algorithm}
\usepackage[noend]{algpseudocode}
\usepackage{color}
\usepackage{array}
\usepackage{subfigure}
\begin{document}

\markboth{Sumanth Chennupati, Sai Prasad Nooka}{Adaptive Hierarchical Decomposition of Large Deep Networks}

%%%%%%%%%%%%%%%%%%%%% Publisher's area please ignore %%%%%%%%%%%%%%%
%
% \catchline{}{}{}{}{}
%
%%%%%%%%%%%%%%%%%%%%%%%%%%%%%%%%%%%%%%%%%%%%%%%%%%%%%%%%%%%%%%%%%%%%
\title{Adaptive Hierarchical Decomposition of Large Deep Networks}
%\title{Instructions for Typesetting Camera-Ready \\
%Manuscripts using \TeX\ or \LaTeX\footnote{For the title, try not to use
%more than 3 lines. Typeset the title in 10 pt Roman, boldface with the first letter of important words capitalized.}}

\author{Sumanth Chennupati\footnote{Equal Contribution}, Sai Prasad Nooka*, Shagan Sah, Raymond Ptucha}
\address{ Rochester Institute of Technology, Rochester, New York 14623,USA\\\
\email{\{cv7148,spn8235,sxs4337,rwpeec\}@rit.edu}}

\maketitle

%begin{history}
%received{(Day Month Year)}
%revised{(Day Month Year)}
%accepted{(Day Month Year)}
%comby{(xxxxxxxxxx)}
%end{history}

\begin{abstract}
Deep learning has recently demonstrated its ability to rival the human brain for visual object recognition. As datasets get larger, a natural question to ask is if existing deep learning architectures can be extended to handle the 50+K classes thought to be perceptible by a typical human. Most deep learning architectures concentrate on splitting diverse categories, while ignoring the similarities amongst them. This paper introduces a framework that automatically analyzes and configures a family of smaller deep networks as a replacement to a singular, larger network. Class similarities guide the creation of a family from course to fine classifiers which solve categorical problems more effectively than a single large classifier. The resulting smaller networks are highly scalable, parallel and more practical to train, and achieve higher classification accuracy. This paper also proposes a method to adaptively select the configuration of the hierarchical family of classifiers using linkage statistics from overall and sub-classification confusion matrices. Depending on the number of classes and the complexity of the problem, a deep learning model is selected and the complexity is determined. Numerous experiments on network classes, layers, and architecture configurations validate our results.
\end{abstract}


\section{Introduction}

\label{sect:intro}  % \label{} allows reference to this section
We carry in our heads a marvel of the universe, tuned by evolution over millions of years.  In each hemisphere of our brain, the primary visual cortex contains $140$ million neurons, each connecting with up to ten thousand other neurons, forming a massive network with tens of billions of connections. And yet, human vision involves not just $V1$, but an entire series of visual cortices - $V2$, $V3$, $V4$, and $V5$ – each doing progressively more complex visual processing. Recognizing objects isn't easy. Rather, we humans are stupendously, astoundingly good at making sense of what our eyes show us. 

Teaching computers how to perform such object recognition is a challenging task. Convolutional Neural Networks ($CNNs$) offer solutions to such highly complex object recognition challenges and have revolutionized the fields of computer vision and pattern recognition. With respect to object detection, $CNNs$ have demonstrated extraordinary performance on the $1000$-class ImageNet~\cite{deng2009imagenet} dataset, recently surpassing human-level performance~\cite{he2015delving,szegedy2016inception}. Although $CNNs$ have been trained for upwards of $10K$ classes, the number of weights in the fully connected layers grow exponentially, demanding a daunting number of training samples, and consuming huge computational resources.

Deep architectures \cite{krizhevsky2012imagenet,Simonyan2015VeryRecognition,Szegedy2015GoingConvolutions,szegedy2016inception,he2015deep} with hierarchical frameworks enable the representation of complex concepts with fewer nodes than shallow architectures. AlexNet\cite{krizhevsky2012imagenet}, VGG\cite{Simonyan2015VeryRecognition} and GoogLeNet\cite{Szegedy2015GoingConvolutions} are large, deep convolutional neural networks, trained on $1.2$ million high-resolution images and $1000$ classes in the ImageNet Large Scale Visual Recognition Competition ( ILSVRC $2010$). It has been shown that network depth is more important than the number of nodes in each layer\cite{mohamed2012acoustic}, with modern architectures containing over $100$ layers\cite{szegedy2016inception,he2015deep}, requiring the solution of over $100M$ parameters.

As the classification task becomes more difficult, the number of parameters increases exponentially, making the classification not only difficult to train but more likely to overfit the data. It is clear that if number of parameters in a network is too large then the network will try to memorize the patterns rather than trying to generalize the training data. Techniques such as domain adaptation\cite{hoffman2014lsda} and hashing\cite{dean2013fast} have been used to tackle problems with larger classes. Yan et al.\cite{Yan2014HD-CNN:Recognition} leveraged the hierarchical structure of categories by embedding $CNNs$ into category hierarchies.

This paper introduces a multi-layer hierarchical framework to reduce the overall number of solvable parameters by subdividing the classification task into smaller intrinsic problems. Abstract higher level networks initially determine which subnetwork a sample should be directed to, and lower level networks take on the task of finding discriminating features amongst similar classes.  Each sub-network is called a class assignment classifier and can recursively be split into subsequently smaller classifiers. Outputs from these class assignment classifiers predict a test sample’s final class.

Confusion matrices infer class-wise linkage statistics by converting from similarity to dissimilarity matrices. Similarly, k-means and spectral clustering on low dimensional representations of the data offer clues to natural boundaries at a coarser level. By viewing the resulting graph tree, such as a dendrogram graph, logical cluster boundaries can often be determined by manual inspection.  Data driven heuristics along with an iterative search algorithm can automatically detect cluster boundaries. These statistics form a hierarchical representation where classes with different superclass labels exhibit dissimilar features at a higher level and classes with same superclass label (i.e, subclasses) share similar features at lower level. These subclasses belong to a cluster that feed an independent class assignment classifier. To ensure robustness and improved generalization, classes which share similarities across different clusters are encouraged to have multiple superclass labels. Semantic outputs from the activated class assignment classifiers include softmax probabilities. The outputs of the classifiers, feed a final classification engine, which makes the final class decision.

The size and type of architecture of a deep network can have a profound impact on both network accuracy and training resources. Considering the importance of appropriate network configuration for classification tasks, we propose an adaptive network selection approach. The clusters generated during hierarchical clustering exhibit different properties. Classes in these clusters have different statistical characteristics and levels of confusion. Adaptive network configurations should examine the confusion among the subclasses and decide an appropriate network configuration to optimize these individually.

In addition to introducing automated methods for automatically choosing adaptive network configurations, we demonstrate the advantages of using transfer learning as an initialization for the class assignment classifier rather than training an entire network from scratch. Adaptive models determine whether it is best to use pre-trained networks as an initialization or as a fixed feature extractor for each of the families of networks. 

In this paper we make the following contributions: ($1$) a framework that automatically analyzes and configures a family of smaller deep networks as a replacement to a singular, larger network, ($2$) resulting smaller networks are not only highly scalable, parallel and more practical to train, but also achieve higher classification accuracy, ($3$) a method to adaptively select the configuration of the hierarchical family of classifiers.

\section{Background}

The pioneering work of Hubel and Wiesel\cite{TJP:TJP19681951215} laid the foundation for the modern hierarchical understanding of the ventral stream of the primate visual cortex.  Simple receptive fields in the eye form complex cells in V1, then more abstract representations in V2 through V4, and finally into the inferior temporal (IT) cortex.  The object representation in the IT cortex is amazingly robust to position, scale, occlusions, and background- the exact understanding of which still remains a mystery and marvel of the human brain\cite{kruger2013deep}.

Traditional computer vision techniques pair hand crafted low level features such as SIFT\cite{lowe2004distinctive}, SURF\cite{bay2008speeded}, or HOG\cite{dalal2005histograms} along with complimentary classifiers such as support vector machines (SVM) or neural networks.  LeCun et al.\cite{LeCun1998GradientRecognition} introduced convolutional neural networks ($CNNs$), computer vision oriented deep feed forward networks based upon a hierarchy of abstract layers.  $CNNs$ are end-to-end models, learning the low level features and classifier simultaneously in a supervised fashion, giving substantial advantage over methods using independent vision features and classifiers.

Datasets such as MNIST \cite{LeCun1998GradientRecognition}, CalTech\cite{fei2007learning}, and Pascal\cite{zisserman2010pascal} have become more challenging over the years.  The ImageNet\cite{deng2009imagenet} dataset has over $14$M images from over $20,000$ categories.  In $2012$, Krizhevsky and Hinton\cite{krizhevsky2012imagenet} beat the nearest competitor by $10\%$ in the ImageNet Large-Scale Visual Recognition Challenge (ILSVRC)\cite{russakovskyimagenet} competition with a seven layer deep $CNN$, taking advantage of a powerful regularization scheme called dropout\cite{srivastava2014dropout}.

Recent progress in classification accuracy can be attributed to advances in building deeper architectures and improved regularization methods\cite{srivastava2014dropout,goodfellow2013maxout,wan2013regularization,hinton2012improving}. Zeiler \& Fergus\cite{zeiler2014visualizing} improved classification results by introducing random crops on training samples and improved parameter tuning methodologies. Simonyan and Zisserman \cite{Simonyan2015VeryRecognition} investigated the usage of network depth and Szegedy, et al.\cite{Szegedy2015GoingConvolutions} used banks of smaller convolutional filters to simultaneously improve accuracy while decreasing the number of parameters. Zhang, et al.\cite{he2015spatial} computed the feature maps from the entire image only once, and then pooled features in arbitrary regions (sub-images) to generate fixed-length representations.

Other advances are attributed to the use of new non-linear activations \cite{goodfellow2013maxout,maas2013rectifier,nair2010rectified,lin2013network,srivastava2013compete} such as Rectifier Linear Units (ReLU). Zhang, et al. He, et al.\cite{he2015delving} used a parameterized version of ReLU to simultaneously learn slope parameters along with weight hyper parameters during backpropagation. 

There are numerous works describing hierarchical decomposition of classification problems\cite{Tousch2012333}.  One of the earliest attempts of a $CNN$ hierarchical approach\cite{NIPS2013_5029} used transfer learning from clusters with many samples to clusters with few.  Deng et al.\cite{Deng2014} used a hierarchy of label relations, and further improvements were made by\cite{Xiao:2014:EIL:2647868.2654926,Yan2014HD-CNN:Recognition} using two and many categories, respectively. 

To form class clusters, confusion matrices can be used to determine hierarchical clusters \cite{Godbole2002ExploitingClassifiers,podolak2008hierarchical,xiong2012building} increased robustness by allowing classes to fork in more than one hierarchal branch.  Salakhutdinov et al. Salakhutdinov, et al. \cite{salakhutdinov2013learning} combined structured hierarchical Bayesian models with deep learning to generate a framework that can learn new concepts with a minimal number of training samples.  $CNN$ hierarchical improvements were demonstrated by\cite{Szegedy2015GoingConvolutions,howard2013some} and a category hierarchical $CNN$ based classifier, $HD-CNN$, was demonstrated in \cite{Yan2014HD-CNN:Recognition} but the memory footprint and time constraints were the major challenges. In \cite{Yan2014HD-CNN:Recognition} the synsets of ImageNet are used for the coarse category taxonomy. These coarse categories or the taxonomy is automatically built using spectral clustering or linkage statistics. The $HD-CNNs$ in \cite{Yan2014HD-CNN:Recognition}  used pre-training for individual classifiers and fine tune them with end to end training. During testing, $HD-CNN$ uses all classifiers to make a final class estimation.  This requires more computations and memory compared to our proposed model which requires only a single classifier to make the final class estimation.

\section{Methods}

\begin{figure}
\begin{center}
\begin{tabular}{c}
\includegraphics[height=6cm]{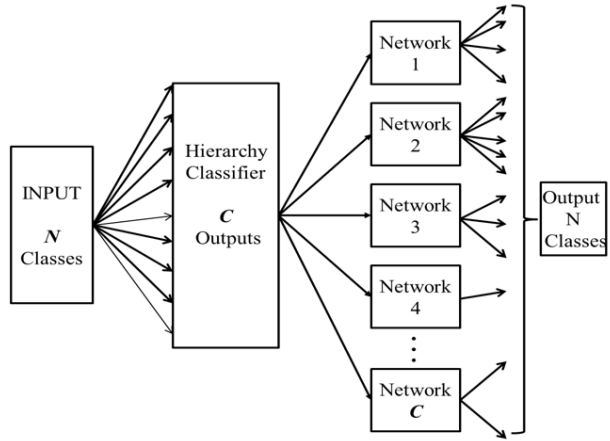}
\end{tabular}
\end{center}
\caption 
{ \label{fig:Fig_3}
Illustration of hierarchical deep network framework.} 
\end{figure}

\begin{figure}[!h]
\begin{center}
\begin{tabular}{c}
\includegraphics[height=6cm]{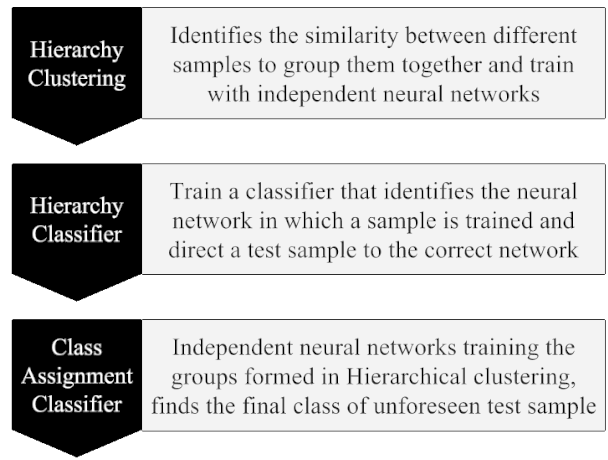}
\end{tabular}
\end{center}
\caption 
{ \label{fig:Fig_1}
Flow of classification using hierarchical deep networks.} 
\end{figure}

We propose a novel method to alleviate the computational complexity involved in training larger networks on datasets with higher number of discrete classes or concepts. Our approach (shown in Fig.~\ref{fig:Fig_3}) uses a high-level classifier to initially determine which cluster a sample belongs to, and then passes that sample into the corresponding class assignment classifier to make the final class assignment.  Moreover, the optimal number of clusters are automatically determined and each subclass in trained independently.  The first stage is hierarchy clustering for determining the number of clusters. This exploits the rich information from a class-to-class confusion matrix (generated using a simplified conventional neural network mapping to all classes or concepts) to find hidden correlations amongst classes and form clusters. After the initial training, the hierarchy classifier predicts a cluster for each sample. This sample is passed into one of $C$ smaller class assignment classifiers, each of which is only concerned with a subset of classes to make a final classification estimate. The approach can be divided into three phases (shown in Fig.~\ref{fig:Fig_1}): (1) hierarchical clustering, (2) hierarchy classifier, and (3) class assignment classifiers. In the following sub-section we will describe each of these modules in detail.

\subsection{Hierarchy Clustering}
To address classification problems with very large number of classes, we propose a hierarchical approach for clustering similar classes into clusters. This requires training of a handful of much simpler neural networks with a reduced number of overall parameters. The intuition for using a hierarchical clustering is the presence of coarse categories or super classes that comprise of a higher number of fine classes. To generate clusters from a given set of classes, we perform spectral clustering on the confusion matrix. The main challenge with such a hierarchical clustering scheme is the selection of an optimum merge or split breakpoints, which if done improperly, can lead to low quality clusters. To address this issue, we formulate an automated multi-phase technique that is based on the analysis of the class confusion matrix of the classifier in the parent stage.

We use the linkage statistics from a class confusion matrix for getting correlation indicators among classes in a hierarchical configuration. The distance matrix ($D$) is estimated from the class confusion matrix ($C$), and measures the dissimilarity among different classes. $D$ is computed by a simple three step process: (i) transform similarity to dis-similarity $D = 1 - C$, (ii) set self dis-similarity to zero $D_{ii} = 0$, and (iii) make the matrix symmetric $D = 0.5 * (D + D^T)$. In this paper, $C_{i}$ represents class $i$ and $Q_{i}$ represents cluster $i$. $D_{p}$ ($C_{i}, C_{j}$) represents the dissimilarity between Classes $C_{i}$ and $C_{j}$ and $D_{p}$ ($Q_{i}, Q_{j}$) represents the dissimilarity between clusters $Q_{i}$ and $Q_{j}$.

In order to create coarse and fine categories (shown in Fig. \ref{fig:Fig_2}), a multi-stage iterative process is deployed to sub-divide the parent cluster into smaller class clusters until a termination criterion is met. In a stage $p$, $D$ will have dimensions $K_{p} \times K_{p}$, where an element $D_{p}$ ($Q_{i}, Q_{j}$) represents the dissimilarity between clusters $i$ and $j$. An unweighted pair cluster method based on the arithmetic mean is used for determining the linkages between individual clusters. $D_{p}$ $(Q_{i}, Q_{i}) = 0 \hspace{2mm}  \forall\hspace{1mm} i\hspace{1mm} \in K$, represents the dissimilarity of a cluster with itself. We use a top-down divisive strategy to find non-overlapping classes that starts by including all classes in a single cluster. The dissimilarity between clusters dynamically determines the split points with an upper limit on the number of classes in a cluster. As a result, this technique automatically adapts to the internal characteristics of the data. This is one of the most important advantages with using such an approach that adapts to the dataset.

Small non-overlapping clusters are obtained by clustering similar classes together. However, during test time in a non-overlapping setting, a misclassified sample at a parent level would always be predicted incorrectly at the lower levels. Therefore to achieve a higher generalization, classes in smaller clusters are overlapped using the posterior probabilities. The confusion matrix of the parent cluster are column normalized  $DCN_{p}$  to obtain the class posterior probabilities. An element $DCN_{p}$ ($C_{i}, C_{j}$) represents the likelihood that a sample is of the true class $C_{i}$ given that it was predicted as class $C_{j}$. Let cluster $Q_{i}$ be the collection of classes $C_{i, ... ,n}$ , then the condition that a certain class $C_{j}$ is similar to this cluster $Q_{i}$ can be given as, 
\begin{equation}
\label{eq:1}
DCN_{p}(C_{i},C_{j})\geq (\gamma.K_{p-1})^{-1} ~ \forall C_{i} \in Q_{i}, C_{j} \notin Q_{i} 
\end{equation}

\begin{figure}[!h]
\centering
\subfigure[ Example of a dendrogram with dissimilarity among the classes.]{
\centering
     \includegraphics[trim= 0cm 15cm 0cm 0cm, clip=true, totalheight=0.35\textheight,angle=0]{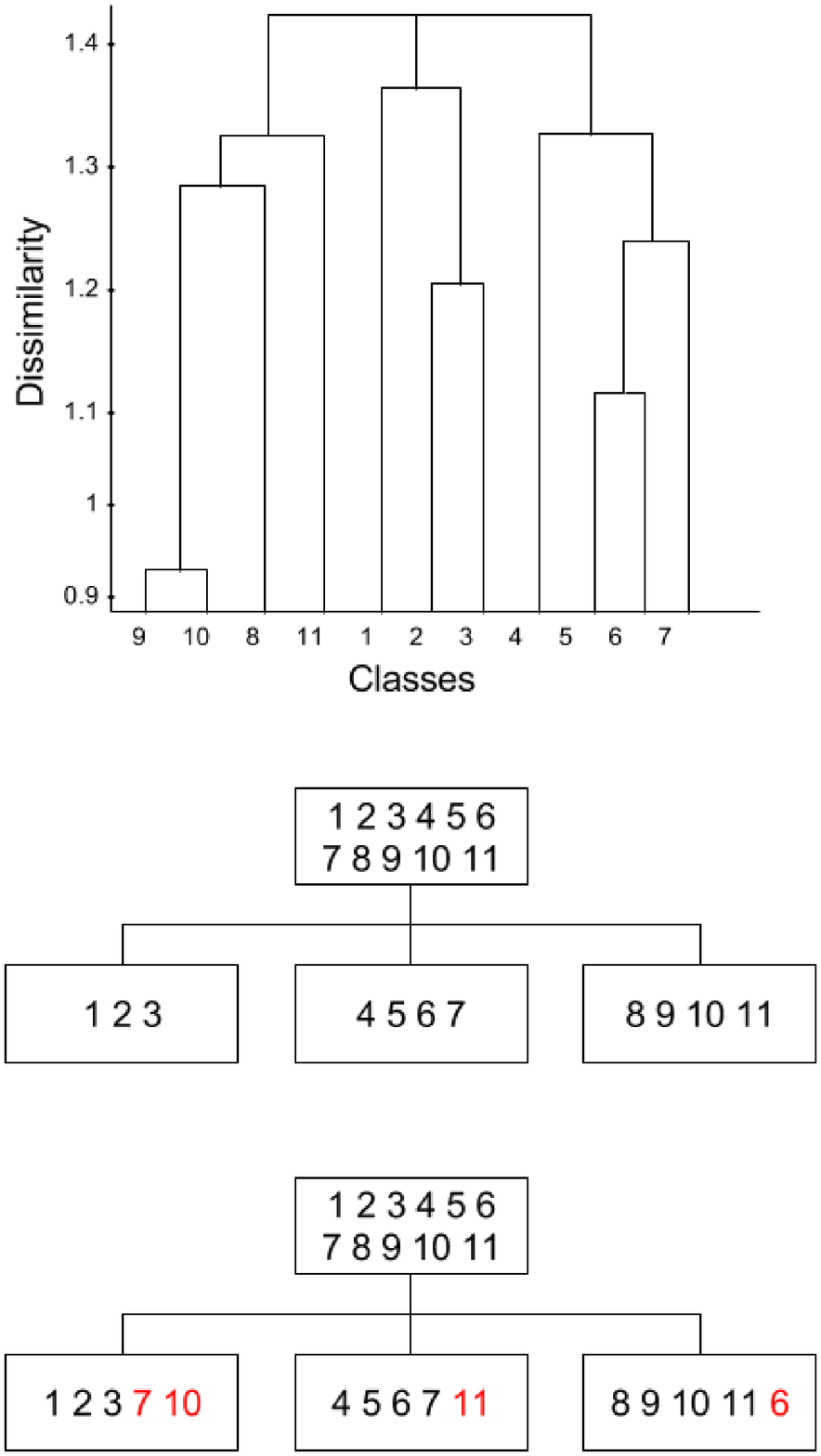}
}
\subfigure[ Example of the non-overlapping clusters formed.]{
\centering
     \includegraphics[trim= 0cm 6.5cm 0cm 14cm, clip=true, totalheight=0.15\textheight,angle=0]{Fig_2.eps}
}
\subfigure[ Example of the overlapping clusters.]{
\centering
     \includegraphics[trim= 0cm 0cm 0cm 21.5cm, clip=true, totalheight=0.15\textheight,angle=0]{Fig_2.eps}
}
\caption[Optional caption for list of figures]{Illustration of hierarchy clustering on Toy data having 11 classes. }
\label{fig:Fig_2}
\end{figure}
We use a parametric threshold of ($\gamma.K_{p-1})^{-1}$, where $\gamma$ is an overlapping hyper-parameter that determines the probability for including a class $C_{j}$ in cluster $Q_{i}$. The value of $\gamma$ depends on the number of classes in the original problem and the number of clusters in the parent stage. Since the clustering is done in a hierarchical fashion, the threshold is dependent on the number of classes in the previous stage $(K_{p-1})$. Thus, all classes that are not a part of cluster $Q_i$ are compared with the cluster. The overlapping in the classifier allows a test sample to follow multiple paths of sub-classifiers. Algorithm~\ref{Hcc} describes the pseudo code of the class clustering algorithm.

\makeatletter
\def\BState{\State\hskip-\ALG@thistlm}
\makeatother
\begin{algorithm}
\caption{Hierarchy Class Clustering}\label{Hcc}
Hierarchy relationships between classes are derived using the confusion matrix $C_{p}$ that measures linkage distances $d$ between classes. To form clusters with overlapping classes, we threshold class posterior probabilities $DCN$ for classes originally not in cluster.

\textbf{Input:} Confusion matrix $C_{p}$ at classification stage $p$

\textbf{Output:} Overlapping class labels $Q$

\textbf{Initialize:} Upper limit on non-overlapping cluster size θ and overlapping factor $\gamma$

\begin{algorithmic}[1]
\State Compute distance matrix $D$ from $C_{p}$
\State Compute linkage statistics:

$d(r,s) = \frac{1}{n_{r}.n_{s}} \sum_{i=1}^{n_{r}}\sum_{i=1}^{n_{s}} dist(x_{r_{i}},x_{s_{j}})$

Where $x_{r_{i}}$ and $x_{s_{j}}$ are dissimilar clusters with $n_{r}$ and $n_{s}$ elements, respectively

\BState Compute cumulative linkage values $Cum(d)$ 

\textbf{for} descending values $k$ in $Cum(d)$

\hspace{1cm}$\alpha = $ no. of classes with $d$ \textless $k$

\hspace{1cm} \textbf{if} $\alpha$ \textgreater $\theta$ \textbf{then}

\hspace{1.5cm} cluster classes $\alpha$ as a new cluster $Q$

\textbf{end}
\BState Compute column normalized confusion matrix $(DCN)$

\textbf{for} each cluster $Q_{i}$

\hspace{1cm} \textbf{if} $DCN_{p}(C_{i},C_{j})\geq (\gamma.K_{p-1})^{-1}  \forall C_{i} \in Q_{i}, C_{j} \notin Q_{i} $ \textbf{then}

\hspace{1.5cm} append class $j$ to cluster $Q_{i}$

\textbf{end}
\end{algorithmic}
\end{algorithm}

\subsection{Hierarchy Classifier}
Let $S$ be the training set with $N$ classes, where $C$ clusters are formed after hierarchical clustering, such that $C$ clusters have $n_{1}, n_{2}, n_{3} ... n_{c}$  number of classes and $n_{1}+ n_{2} + n_{3} + ... + n_{c} =N$. The linkage statistics define a high-level classifier, called the hierarchy classifier, which determines the cluster a sample belongs to. Samples are passed into the corresponding sub-class network to make the final class assignment. These subclass networks are referred to as class assignment classifiers.

\subsection{Class Assignment  Classifier}
The class assignment classifier consists of $C$ smaller neural networks, each predicting a unique and inclusive subset of the $N$ classes. Each of the $C$ class assignment classifiers output their respective subset of $N$ classes, i.e., all classes of the dataset are classified at this stage of hierarchical model. In order to mitigate misclassifications from the previous stage of hierarchy classifier during testing, overlapping clusters allow a sample to pass to more than one class assignment classifier. Let $p_{1}, p_{2}, p_{3}, ... ,p_{c}$ be predictions of the hierarchy classifier for corresponding $C$ outputs and $q_{1}, q_{2}, q_{3}, ... ,q_{c}$ be the predictions of $Network_{1}$ for the corresponding $n_{1}$ outputs. The top $k$ predictions from overlapping clusters are predictions of the hierarchy classifier and the class assignment classifier. The final predicted classification output is referred as confidence $(\mathcal{C})$: 

\begin{equation}
\label{eq:2}
Confidence(\mathcal{C}_{i}) = p_{j}\times q_{k}  \forall i \in (1, C), k \in (1,n_{j}) 
\end{equation}

To boost the performance of hierarchical models, this study introduces an adaptive network selection manager as shown in Fig. 4. The network selection manager considers both the number of classes as well as cluster heuristics to select an appropriate network configuration for each class assignment classifier. Experiments determine the best configuration for a given cluster. A CNN with a preset number of layers and filters is used as a base configuration and the adaptive network model determines the configuration used for any cluster.
\begin{figure}[!h]
\centering
\subfigure{
\centering
     \includegraphics[trim= 0cm 0cm 0cm 0cm, clip=true, totalheight=0.3\textheight,angle=0]{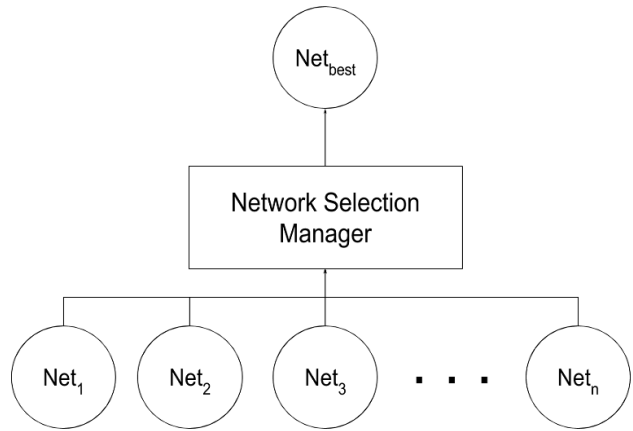}
}
\caption[Optional caption for list of figures]{Illustration of a network selection manager that chose between n dierent networks. }
\label{fig:Fig_2}
\end{figure}
\begin{figure}[!h]
\begin{center}
\begin{tabular}{c}
\includegraphics[height=10cm]{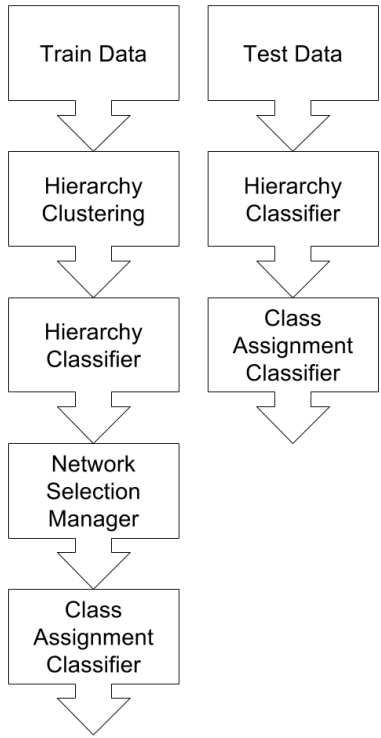}
\end{tabular}
\end{center}
\caption 
{ \label{fig:Fig_5}
Illustration of updated hierarchical model with Network selection manager.} 
\end{figure}

Fig.~\ref{fig:Fig_5} shows the stages of an adaptive hierarchical model. It includes a network selection manager that selects the best network configuration for the class assignment classifier.  During training, the hierarchy classifier predicts a cluster a sample belongs and passes it through one of $C$ smaller class assignment classifiers that are chosen by the network selection manager.

The following sections will discuss the adaptive network selection strategy and the adaptive transfer learning variants of the network selection manager.

\subsection{Adaptive Network Selection}
 \label{adaptive_network_selection}
The number of layers in the network and number of nodes per layer are both important in optimizing classification problems. Since hierarchical clusters exhibit unique statistical properties, selecting an appropriate network would typically improve the overall classification accuracy. Attributes such as the number of classes, confusion matrix linkage and class-to-class correlation statistics allow an adaptive network framework to make the classifier decision. It selects a network from a set of pre-configured CNN architectures shown in  Tables 1 \& 2 which vary in number of layers and nodes/filters per layer. Rather, for Multi-Layer Perceptrons $(MLPs)$ the decision on network architecture- layers and number of nodes, for a cluster is treated as a regression problem.

To train such a regressive model, multiple network configurations were trained on different clusters, where each cluster has its unique statistical properties. The model learns to correlate cluster properties with the corresponding best network architecture. This regression model provides 2 outputs for MLPs : 1) number of layers required, 2) Number of neurons required in each layer.

\begin{algorithm}
\caption{Adaptive Network Selection}\label{ANS}
\vspace{2mm}
The network selection manager outputs the best network configuration for a class assignment classifier based on cluster heuristics.

%\vspace{2mm}
\textbf{Input:} Non- Overlapping class labels $C$ or Overlapping class labels $Q$

%\vspace{1.5mm}
\textbf{Output:} Best network configuration $Net_{best}$

%\vspace{1.5mm}
\textbf{Initialize:} $Net_{best} = Net_{1}$

%\vspace{1.5mm}
\begin{algorithmic}[1]
\State Send this cluster into pre-defined classifier and generate confusion matrix $CM$.

%\vspace{1mm}
\State Change the number of nodes in the final layer of the pre-defined networks $(Net_{1}, Net_{2}, Net_{3}, ... ,Net_{n} )$ to classify all subclasses in $C$ or $Q$.

%\vspace{1mm}
\BState Evaluate accuracy $\mathcal{A}_{i}$ over all samples that contain class labels $C$ or $Q$. 

\textbf{for} all networks $Net_{i}$ in $Net_{1}, Net_{2}, Net_{3}, ... , Net_{n}$

\hspace{1cm}$\mathcal{A}_{i}$ = Accuracy of network $Net_{i}$ for all samples with subclass labels in $C$ or $Q$.  

\hspace{1cm} \textbf{if} $\mathcal{A}_{i}$ \textgreater  $\mathcal{A}_{best}$  \textbf{then}

\hspace{1.5cm} $Net_{best}$ = $Net_{i}$

\hspace{1cm} repeat until all networks are evaluated.

\textbf{end}

%\vspace{1mm}
\BState Pile up the dataset using the best network as label and features as confusion matrix $CM$
%\vspace{1mm}
\BState Depending upon MLPs or CNNs used, regression or classification networks are trained using the above generated dataset. 
\end{algorithmic}
\end{algorithm}

\subsection{Adaptive Transfer Learning}

CNNs learn to extract features along with a feature classifier for unsurpassed image classification performance. It has also been widely shown that transfer learning enable both higher classification accuracies and faster convergence when applied to new classification problems. For example, the popular architectures of AlexNet\cite{krizhevsky2012imagenet}, VGG\cite{Simonyan2015VeryRecognition}, and GoogLeNet\cite{Szegedy2015GoingConvolutions} that are pre-trained on ImageNet offer an excellent weight initialization for new visual classification problems. The first few layers of CNNs describe high level abstract features which apply equally to most visualization problems, negating the need for further fine-tuning. Initializing all weights from a previously learned network and fine-tuning last few layers typically offers increased performance. Furthermore, such an initialization requires only a few training epochs are required for converge. Experiments demonstrate the impact that transfer learning has on the adaptive architectural selection. 

\subsection{Joint Model}
In earlier sections, either convolutional neural networks or multi-layer perceptron neural networks in an adaptive mode are used. We now allow this selection in conjunction to select the optimum model configuration for a particular sub-network.

\section{Experiments \& Results}
Several experiments were performed to demonstrate the application of the hierarchical models and adaptive hierarchical models. Multiple architectures were carefully pre-defined to feed the network selection manager which selects the best possible network for different class assignment classifiers. The CalTech-$101$, CalTech-$256$, CIFAR-$100$, ImageNet$100$, ImageNet$200$ and ImageNet datasets were used. The CalTech-$101$dataset has $102$ classes, with $40$ to $800$ images per class with image size of $300\times 200\times 3$ pixels. The CalTech-$256$ dataset has $257$ classes, with $80$ to $827$ images per class with image size of $300\times 200\times 3$ pixels. These images were resized to $64\times 64$. Training and testing splits for CalTech datasets were generated using 6-fold cross validation. The CIFAR-$100$ dataset has $100$ classes with 500 images for training and $100$ images for testing respectively per class and has an image size of $32\times 32\times 3$ pixels. CIFAR-$100$ consists of $100$ fine categories and $20$ coarse categories with $32\times 32$ RGB images. ImageNet dataset has $1000$ classes with $1350$ samples per class and images are resampled to $224\times 224$ RGB images for all experiments in this paper. ImageNet$100$ and ImageNet$200$ are subsets of ImageNet with $100$ and $200$ classes randomly selected from the $1000$ classes.

\subsection{Training}

For all experiments in this paper baseline CNN configuration refers to $CNN_{1}$ from the pre-defined configurations described in Table 1. For all transfer learning experiments $CNN_{TL_{1}}$ refers to baseline. The pre-defined CNNs, $CNN_{2}$ through $CNN_{5}$, each contain six layers, the first four being convolution layers followed by pooling and the final two are fully connected layers with a dropout ratio of $0.5$. Conv $5\times5|32$ Stride:$4$ indicates convolutional operation with $32$ filters of size of $5\times5$ with a stride factor $4$. ReLU, Maxpool$|2$ allows convolutional output to non-linear activation function followed by maxpooling with stride $2$. $FC$ is the number of nodes in the fully connected layer. All networks are trained for $40$ epochs with a learning rate $0.01$ and momentum $0.9$. The adaptive transfer learning configurations, $CNN_{TL_{2}}$ through $CNN_{TL_{5}}$, each contain 7 weight layers of which the first five are convolution layers and rest are fully connected layers with the same dropout ratio of $0.5$. $CNN_{TL_{1}}$ is identical to the VGG-f configuration \cite{Simonyan2015VeryRecognition}. 

\begin{table}[th]
\tbl{Pre-defined convolutional neural network configurations for CalTech-$101$, CalTech-$256$.}
{\begin{tabular}{@{}p{1cm}p{1.5cm}p{1.5cm}p{1.5cm}p{1.5cm}p{1.5cm}@{}} \toprule
{Net} & {\centering \footnotesize $CNN_{1}$} &{\centering \footnotesize $CNN_{2}$} & {\centering \footnotesize$CNN_{3}$} &{\centering \footnotesize $CNN_{4}$} & {\centering \footnotesize$CNN_{5}$}
\\ \colrule

Depth & $\centering 6$ & $\centering 6$ & $\centering 6$ & $ \centering 8$ & $\centering 10$  
\\

 \multicolumn{6}{c}{ Input:$64 \times 64$ RGB image}  
\\ 
Conv &  $5\times 5|32$ &  $3\times 3|32$ & $7\times 7|32$ &  $5\times 5|32$ &  $5\times 5|32$  
\\ 
 \multicolumn{6}{c}{ReLU,Maxpool$|2$} 
\\
Conv &  $5\times 5|32$ &  $3\times 3|32$ & $5\times 5|32$ &  $3\times 3|32$ &  $3\times 3|32$ 
\\
 &   &   &  &   &  $3\times 3|32$  
\\ 
 \multicolumn{6}{c}{ReLU,Maxpool$|2$} 
\\
Conv &  $5\times 5|64$ &  $3\times 3|64$ & $3\times 3|64$ &  $3\times 3|32$ &  $3\times 3|32$ 
\\
 &   &   &  &   $3\times 3|64$ & $3\times 3|64$ 
\\
 \multicolumn{6}{c}{ReLU,Maxpool$|2$} 
\\
Conv &  $5\times 5|64$ &  $3\times 3|64$ & $3\times 3|64$ &  $3\times 3|64$ &  $3\times 3|64$ 
\\
 &   &   &  &   $3\times 3|64$ & $3\times 3|128$ 
\\ 
 \multicolumn{6}{c}{ReLU,Maxpool$|2$} 
\\
FC &  1024 &  512 & 256 &  512 &  512
\\ 
 \multicolumn{6}{c}{ReLU, Dropout} 
\\
FC &  512 &  256 & 128 &  128 &  256 
\\
& & & & & 128  
\\ 
 \multicolumn{6}{c}{ReLU, Dropout} 
\\
FC & C &  C & C &  C &  C 
\\ \botrule

\end{tabular}}
\end{table}

\subsubsection{Hierarchical Model}

The hierarchical model consists of a hierarchical classifier which makes coarse category predictions and a class assignment classifier that predicts the final class category of a test image. In a simple hierarchical model, a sample with its coarse category label is sent to a hierarchical classifier which learns the coarse category representations present in the dataset. Later, the sample is sent to one of the class assignment classifiers based on the fine category label. This allows the hierarchical model to learn the coarse and fine category representations in the dataset and effectively use them to make final predictions. 

In the MLP experiments, a single large network contains $2$ hidden layers  with dimensions $[200,150]$ and a hierarchical classifier, while class assignment classifiers use a network with $2$ hidden layers of size $[25,10]$. For experiments with the CNNs, networks described in Table 1. $CNN_{2}$ and $CNN_{3}$ are used as hierarchical and class assignment classifiers respectively for CalTech-$101$ and CalTech-$256$ datasets. For CIFAR-$100$, the last layer of all the networks is replaced to fit the number of classes.
% All the networks ($MLPs$ and $CNNs$) are trained with a random weight initialization for $40$ epochs with a learning rate $0.01$ and momentum $0.9$.

\subsubsection{Adaptive Hierarchical Model}

Adaptive hierarchical model slightly differs from the hierarchical model in the second stage of classification where a sample is processed for fine category prediction. In the training phase, a sample directed towards a class assignment classifier is sent to all pre-defined networks and the best model is selected by the network selection manager that is later used in the testing phase to make fine category predictions.

In experiments with MLPs, several combinations of networks were used as described in section \ref{adaptive_network_selection} and the resulting regression model was used to predict the class assignment classifier for the testing phase. In experiments with CNNs, information from $CNN_{2}$ through $CNN_{5}$ was used by the network selection manager to select the best network configuration for a given cluster.

\subsubsection{Adaptive Transfer Learning }

In the hierarchical experiments on ImageNet dataset, we used $CNN_{TL_{1}}$ as the base model, $CNN_{TL_{3}}$ for making coarse category predictions and $CNN_{TL_{5}}$ for making fine category predictions. For adaptive transfer learning, adaptive network section manager chooses between $CNN_{TL_{2}}$ through $CNN_{TL_{5}}$ from Table 2.  Layers in blue are are optimized by fine tuning and all other layers are fixed. Pre-trained weights for transfer learning experiments are obtained from VGG\cite{Simonyan2015VeryRecognition} that was trained on ImageNet dataset for $20$ epochs.

\begin{table}[th]
\tbl{Pre-trained configurations for ImageNet datasets.}
{\begin{tabular}{@{}p{1cm}p{1.5cm}p{1.5cm}p{1.5cm}p{1.5cm}p{1.5cm}@{}} \toprule
{Net}  &{\centering \footnotesize $CNN_{TL_{1}}$} & {\centering \footnotesize$CNN_{TL_{2}}$} & {\centering \footnotesize$CNN_{TL_{3}}$} &{\centering \footnotesize $CNN_{TL_{4}}$} &{\centering \footnotesize $CNN_{TL_{5}}$}
\\ \colrule

Depth  & $\centering 8$ & $\centering 8$ & $\centering 8$ & $\centering 8$ & $\centering 8$ 
\\

 & \multicolumn{5}{c}{ Input:$224 \times 224$ RGB image} 
\\ 
Conv & $11\times 11|64$& $11\times 11|64$ &  $11\times 11|64$ &  $11\times 11|64$ & $11\times 11|64$ 
\\ 
& Stride:4 & Stride:4 &  Stride:4 &  Stride:4 & Stride:4
\\ 
 & \multicolumn{5}{c}{ReLU,Maxpool$|2$}
\\\
Conv &   $5\times 5|256$& $5\times 5|256$ &  $5\times 5|256$ &  $5\times 5|256$ & $ \color{blue} 5\times 5|256$ 
\\
 & \multicolumn{5}{c}{ReLU,Maxpool$|2$}
\\\
Conv & $3\times 3|256$& $3\times 3|256$ &  $3\times 3|256$ &  $3\times 3|256$ & $ \color{blue} 3\times 3|256$ \\
& \multicolumn{5}{c}{ReLU}
\\
Conv & $3\times 3|256$& $3\times 3|256$ &  $3\times 3|256$ &  $ \color{blue} 3\times 3|256$  & $ \color{blue} 3\times 3|256$ 
\\
 &  $3\times 3|256$& $3\times 3|256$ & $3\times 3|256$ &$ \color{blue} 3\times 3|256$   & $ \color{blue} 3\times 3|256$ 
\\ 
 & \multicolumn{5}{c}{ReLU,Maxpool$|2$}
\\
FC &  4096 & 4096 & $ \color{blue}2048 $& $\color{blue}2048$ &$\color{blue} 1024$  
\\ 
& \multicolumn{5}{c}{ReLU, Dropout}
\\
FC & 4096 & $\color{blue}2048$ &  $\color{blue}2048$ & $\color{blue}1024$ &$\color{blue} 1024$  
  
\\ 
 & \multicolumn{5}{c}{ReLU, Dropout}
\\
FC & $C$ & $\color{blue}C $&  $\color{blue}C$ & $\color{blue}C$ &$\color{blue} C $ 
\\ \botrule

\end{tabular}}
\end{table}

\subsection{Results }

Table 3. demonstrates that a MLP processing on the CalTech-$101$ dataset increases the final accuracy by approximately $16\%$ using a non-overlapping hierarchical architecture. Similar observations were recorded with overlapping hierarchical architecture, but performance decreases with increasing overlap factor. This was attributed to the increase in confusion in the hierarchical stage. It should also be noted that the memory requirements increase as the overlap factor increases due to larger class assignment classifiers.

\begin{table}[th]
\tbl{Performance of hierarchical models built using MLPs on CalTech-101 dataset.}
{\begin{tabular}{@{}ccccc@{}}
\toprule
C & Clustering Method & Gamma($\gamma$) & $HC(\%)$ & $FC(\%)$\\
\colrule
$1$ & NA & NA & NA & $45.6$\\
$44$ & Non-Overlap & NA & $69.43$ & $61.39$\\
$44$& Overlap & $3$ & $69.05$ & $\textbf{61.56}$\\
$44$ & Overlap & $5$ & $62.73$ & $60.13$\\
$44$ & Overlap & $8$ & $52.05$ & $58.61$\\
\botrule
\end{tabular}}\begin{tabnote}
Top line indicates performance of a single large MLP neural network with two hidden layer of dimensions $[200,150]$. MLP neural network with two hidden layers of dimensions $[25,10]$ used in each mini-network for the rest of the models. Hierarchical clustering is controlled by varying the parameter gamma ($\gamma$). $HC$ indicates hierarchy classifier accuracy and $FC$ indicates final classification accuracy.\end{tabnote}

\end{table}

In Table 4., evaluation using CNNs decreased the final accuracy by $4\%$ when using a non-overlapping hierarchical architecture. Potential reasons explaining this decline are $1$) the identical architecture of all the mini-networks, and $2$) when a cluster has fewer number of classes, the number of training samples for that network are also less, making them insufficient for training CNNs.
\begin{table}[!h]
\tbl{Performance of hierarchical models built using CNNs on CalTech-101 dataset.} 
{\begin{tabular}{@{}ccccc@{}}\toprule
C & Clustering Method & Gamma($\gamma$) & $HC(\%)$ & $FC(\%)$\\\colrule
$1$ & NA & NA & NA & $55.84$\\
$48$ & Non-Overlap & NA & $62.42$ & $51.57$\\
$48$ & Overlap & $3$ & $50.33$ & $50.72$\\\botrule
\end{tabular}}
\begin{tabnote}
Top line indicates performance of a single large CNN. Similar CNNs are used in each mini-network of other experiments.\end{tabnote}
\end{table}

\begin{table}[!h]
\tbl{Performance of hierarchical models built using MLPs on CalTech-$256$ dataset.}
{\begin{tabular}{@{}ccccc@{}}\toprule
C & Clustering Method & Gamma($\gamma$) & $HC(\%)$ & $FC(\%)$\\ \colrule
$1$ & NA & NA & NA & $18.61$\\
$104$ & Non-Overlap & NA & $23.07$ & $21.56$\\
$104$ & Overlap & $3$ & $24.49$ & $21.96$\\
$104$ & Overlap & $5$ & $22.55$ & $20.61$\\ \botrule
\end{tabular}}
\end{table} 

\begin{table}[!h]
\tbl{Performance of hierarchical models built using CNNs on CalTech-$256$ dataset} 
{\begin{tabular}{ @{}ccccc@{}} \toprule
C & Clustering Method & Gamma($\gamma$) & $HC(\%)$ & $FC(\%)$\\
$1$ & NA & NA & NA & $36.21$\\ \colrule
$104$ & Non-Overlap & NA & $29.87$ & $28.34$\\
$104$ & Overlap & $3$ & $30.65$ & $25.62$\\
$104$ & Overlap & $5$ & $27.38$ & $21.47$\\ \botrule
\end{tabular}}
\end{table} 

\begin{table}[!h]
\tbl{Performance of hierarchical models built using MLPs on CIFAR-$100$ dataset.} 
{\begin{tabular}{@{}ccccc@{}} \toprule
C & Clustering Method & Gamma($\gamma$) & $HC(\%)$ & $FC(\%)$\\
$1$ & NA & NA & NA & 22.83\\\colrule
$30$ & Non-Overlap & NA & $28.37$ & $24.96$\\
$30$ & Overlap & $3$ & $27.89$ & $24.82$\\
$30$ & Overlap & $5$ & $26.34$ & $23.25$\\ \botrule
\end{tabular}}
\end{table}

\begin{table}[!h]
\tbl{Performance of hierarchical models built using CNNs on CIFAR-$100$ dataset.} 
{\begin{tabular}{@{}ccccc@{}} \toprule
C & Clustering Method & Gamma($\gamma$) & $HC(\%)$ & $FC(\%)$\\
$1$ & NA & NA & NA & $22.83$\\ \colrule
$30$ & Non-Overlap & NA & $28.37$ & $24.96$\\
$30$ & Overlap & $3$& 2$7.89$ & $24.82$\\
$30$ & Overlap & $5$ & $26.34$ & $23.25$\\ \botrule
\end{tabular}}
\end{table}

Table 5 \& Table 7 show that the accuracy increases by $3\%$ in cases of CalTech-$256$ and CIFAR-$100$ datasets when MLP neural network was used to evaluate the performance of the non-overlapping and overlapping hierarchical architectures. Table 6 \& Table 8, show that final accuracy decreased when using CNNs to evaluate CalTech-$256$ and CIFAR-$100$ datasets. This can be attributed to the issue with smaller sized clusters as explained earlier.\\

\begin{figure}[!h]
\centering
\subfigure[Example of a Dendrogram with $102$ Classes of CalTech-$101$ dataset generated using confusion matrix obtained from a single CNN (Better viewed in color).]{
\centering
     \includegraphics[trim= 0cm 15.5cm 0cm 0cm, clip=true, totalheight=0.35\textheight,angle=0]{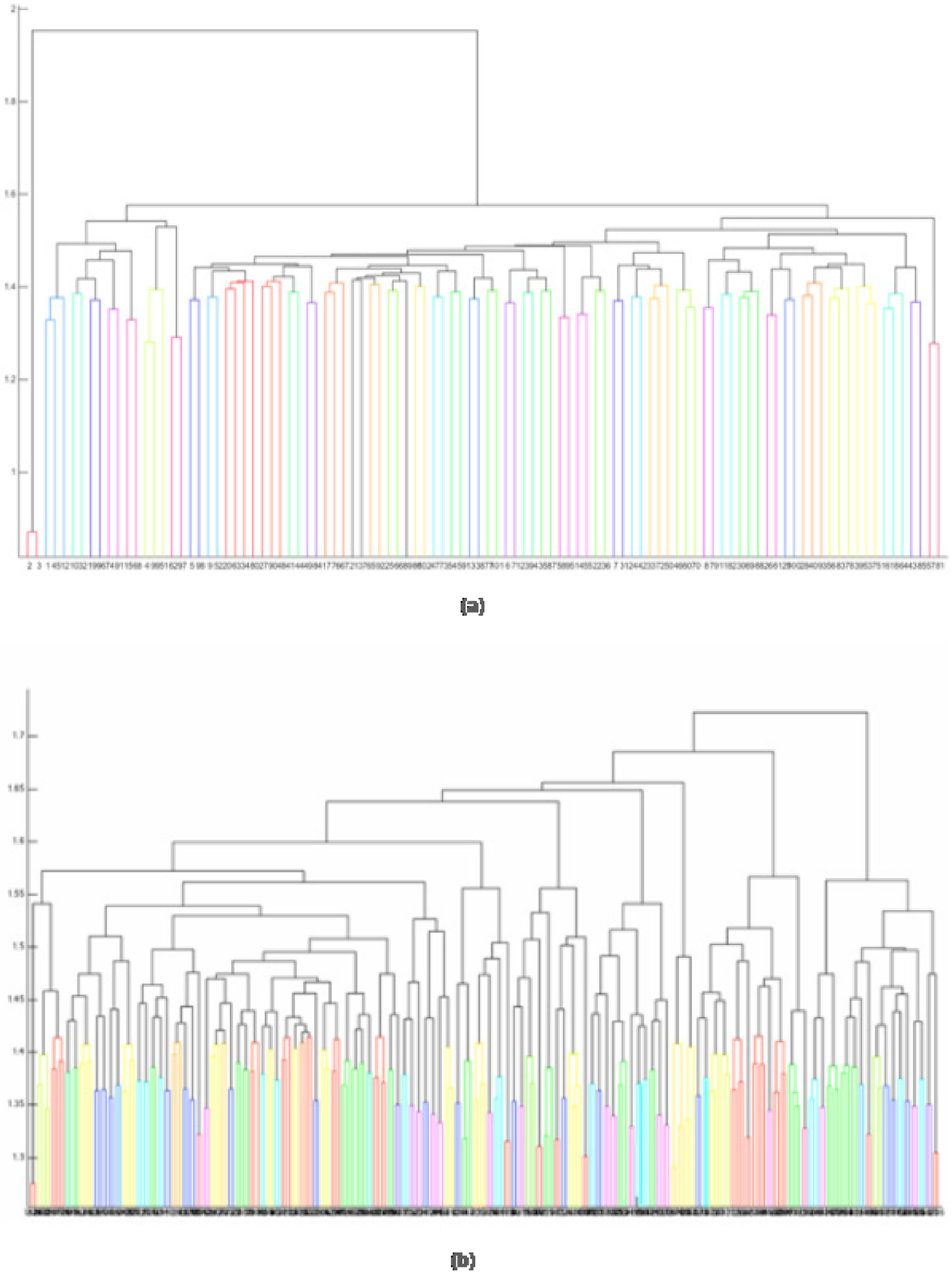}
    \label{fig:Fig_6}
}
\subfigure[Example of a Dendrogram with $257$ Classes of CalTech-$256$ dataset generated using confusion matrix obtained from a single CNN.]{
     \centering
  \includegraphics[trim= 0cm .75cm 0cm 16cm, clip=true, totalheight=0.32\textheight,angle=0]{Fig_6.eps}
    \label{fig:Fig_7}
}
\caption[Optional caption for list of figures]{Examples of  Dendrograms for CalTech Datasets}
\label{fig:subfigureExample}
\end{figure}

\begin{figure}[!h]
\centering
\subfigure[Example of a dendrogram with $100$ Classes of CIFAR-$100$ dataset generated using confusion matrix obtained from a single CNN.]{
\centering
     \includegraphics[trim= 0cm 14cm 0cm 0cm, clip=true, totalheight=0.38\textheight,angle=0]{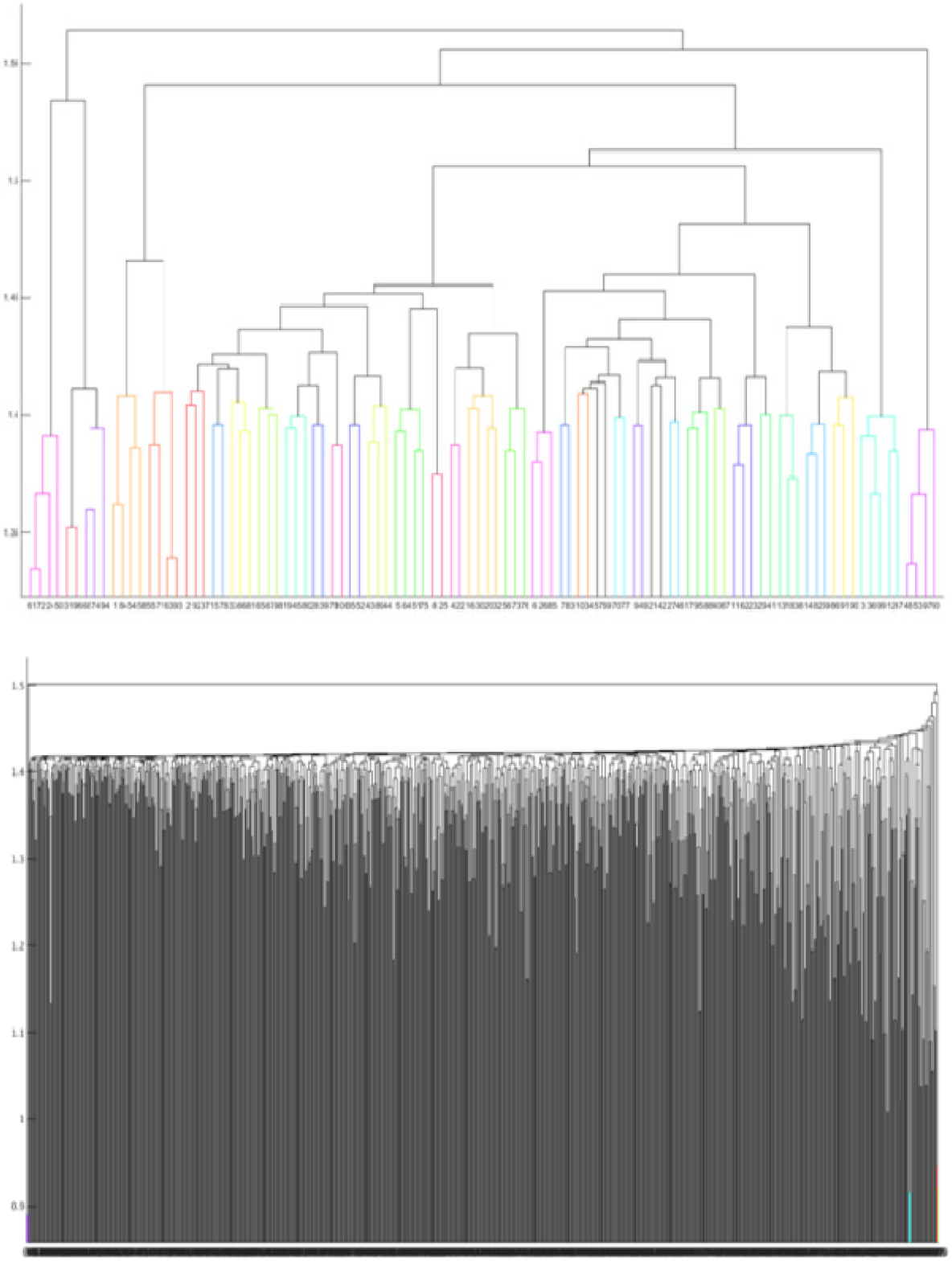}\hfill
   \label{fig:Fig_8}
}
\subfigure[Example of a dendrogram with $1000$ Classes of ImageNet dataset generated using confusion matrix obtained from a single CNN.]{
     \centering
 \includegraphics[trim= 0cm 0.5cm 0cm 14cm, clip=true, totalheight=0.38\textheight,angle=0]{Fig_7.eps}\hfill
 \label{fig:Fig_9}
}

\caption[Optional caption for list of figures]{Examples of a Dendrogram for CIFAR-$100$ and Imagenet datasets.}
\label{fig:subfigureExample}
\end{figure}

The dendrograms in Fig.~\ref{fig:Fig_6},~\ref{fig:Fig_7},~\ref{fig:Fig_8},~\ref{fig:Fig_9}, represent the class clustering formed using the linkage statistics for different data sets. The colors in the graph depict class clusters as determined by the algorithm described in Algorithm.~\ref{Hcc}. The height of the bars represent the magnitude difference between the clusters. It shows the decomposition of the clusters from a top-down perspective on different datasets.  For example, Fig.~\ref{fig:Fig_6} shows that the first two classes on the left are different from the remaining classes of the CalTech-$101$ dataset.The relatively flat joining lines in~\ref{fig:Fig_9} along the top are indicative of a fairly balanced dataset. Further manula analysis of the clusters indicate that the that similar classes were clustered together which proves the efficacy of the hierarchical clustering.

CIFAR-$100$ dataset also provides a higher level taxonomy. So, we compared the performance of hierarchical models with the original (ground truth) clusters in Table 9. The CIFAR hierarchical model and CIFAR adaptive network selection models are generated using the CIFAR-$100$ taxonomy. The learned taxonomy improves $CNN_{1}$ accuracy by $5.62\%$ where as CIFAR-$100$ taxonomy improves $CNN_{1}$ accuracy by $1.28\%$. It was observed that our taxonomy generated more clusters, hence finer categories.

\begin{table}[!h]
\tbl{Performance of a single $CNN$ compared with models generated from adaptive clustering and ground truth CIFAR-$100$ taxonomy.}
{\begin{tabular}{@{}cccc@{}}  \toprule
S.No & Model & Clusters & Accuracy(\%) \\ \colrule
1 & $CNN_{1}$ & NA & $42.29$\\
2 & CIFAR hierarchical model & $20$ & $41.82$\\
3 & CIFAR adaptive network selection & $20$ & $42.83$\\
4 & Hierarchical model & $30$ & $37.56$\\
5 & Adaptive network selection & $30$ & $\textbf{44.67}$\\ \botrule
\end{tabular}}
\end{table}

Table 10. lists results on CalTech-$101$, CalTech-$256$, and CIFAR-$100$ datasets. The non-adaptive hierarchical models and adaptive network selection generally do not improve the performance over a single, large CNN. Adaptive transfer learning experiments improve performance over a single CNN (these were not performed on CIFAR-$100$ due to the smaller size of the images in the dataset). We assume these variations in performance are due to the data-driven nature of CNNs. Both the CalTech datasets have significant variation in number of samples per class, and the results presented in this study were obtained on the entire dataset. Since CIFAR-$100$ has sufficient samples per class, class assignment classifiers can optimize network weights, whereas for CalTech-$101$ and CalTech-$256$, many class assignment classifiers have very few training samples. 

\begin{table}[!h]
\tbl{Performance comparison of a single $CNN$ with hierarchical models.} 
{\begin{tabular}{ @{}cccccc@{}} \toprule
Dataset & $CNN_{1}$ & \parbox[c]{1.5cm}{\centering Hierarchical model} & \parbox[c]{1.5cm}{\centering Adaptive network selection} & \parbox[c]{1.5cm}{\centering Adaptive transfer learning}  &\parbox[c]{1cm}{\centering Joint model} \\ \colrule
CalTech-$101$ & $55.12$ & $51.57$ & $53.25$ & $54.85$ & $\textbf{55.12}$ \\
CalTech-$256$ & $36.21$ & $28.34$ & $29.54$ & $\textbf{36.65}$ & $36.54$\\
CIFAR-$100$ & $42.29$ & $37.56$ & $\textbf{44.67}$ & NA  & $\textbf{44.67}$\\ \botrule
\end{tabular}}
\end{table}

We believe the accuracies would be higher if the number of examples were identical across all classes. To validate this assumption, the performance of the non-adaptive hierarchical model as well as adaptive hierarchical models were evaluated on subsets of the ImageNet dataset.
\begin{table}[!h]
\tbl{Performance comparison of a single CNN with hierarchical model and adaptive transfer learning model on ImageNet dataset.} 
{\begin{tabular}{ @{}cccc@{}}\toprule
Model & Clusters& \parbox[c]{2.5cm}{\centering Accuracy (\%)} \\\colrule
$CNN_{TL_{1}}$ & NA & $55.32$  \\ 
Hieracrchical model & $89$ & $50.27$  \\
Adaptive transfer learning & $89$ & $\textbf{56.06}$ \\ \botrule
\end{tabular}}
\end{table}

Table 11. demonstrates 1.3\% increase in accuracy relatively compared to a single $CNN$. It was observed that some classes in the ImageNet possess extreme similarities and led to formation of clusters with many classes or fine categories. This hinders the performance of hierarchy classifier hurting the overall accuracy. We hypothesize that the hierarchy classifier’s performance would be improved if the resulted clusters had an even distribution of classes across them.

\subsection{Analysis}
In order to make the final class prediction, HD-CNN \cite{Yan2014HD-CNN:Recognition} uses a probabilistic averaging layer that takes inputs from all fine category CNNs as well as coarse category CNNs. HD-CNN achieves a lower error at the expense of increased computational complexity. Our models significantly reduce the computations by making final predictions using just one fine category CNN.

Our work aims at lowering the computational complexity while achieving scalability. Although we underperform compared to HD-CNN, our models perform better than their base-line results. The performances of hierarchical models in this work demonstrate the advantages of various hierarchical architectures while improving  baseline results, but are not intended to compete with state-of-the-art results. Tables 12 \& 13 show the complexity analysis in terms of memory footprint for each network used while testing an image. Table 12. show the memory required by different CNN architectures used in the adaptive transfer learning models to predict the final class. To make comparisons with others, we measure the memory requirements for test mini batch size of 50.

\begin{table}[!h]
\tbl{Comparison of memory footprint between different CNN architectures used in adaptive transfer learning model. }
{\begin{tabular}{@{}cccccc@{}} \toprule
Network & $CNN_{TL_{1}}$ & $CNN_{TL_{2}}$ & $CNN_{TL_{3}}$ &$CNN_{TL_{4}}$ & $CNN_{TL_{5}}$ \\ \colrule
Params Memory (MB) & 232.03 & 169.1 & 97.1 & 84.56 & 48.56 \\ 
Data Memory $\times$ 50 (MB)& 314.5 &  314 & 313 & 312 & 311 \\ 
Total Memory (MB)& 546.53 & 483.1  & 410.1 & 396.56 & 359.56 \\ \botrule 
\end{tabular}}
\end{table}

The baseline VGG-f ($CNN_{TL_{1}}$) model requires $546.53$ mega-bytes memory to make a final class estimation. For similar class estimation, the hierarchical classifier requires ($CNN_{TL_{3}}$) $97.1$ mega-bytes and and class assignment classifier ($CNN_{TL_{5}}$) $48.56$ mega-bytes. The hierarchical model shares the data between both stages and requires $313$ mega-bytes. Hence, overall the hierarchical model would require $97.1 + 48.56 + 313 =  458.66$ mega-bytes memory. In case of adaptive transfer learning, it can use any model from $CNN_{TL_{2}}$ to $CNN_{TL_{5}}$, so it would require $97.1 + 169.1 + 314 = 580.2$ mega-bytes in the worst case scenario. Hence, the hierarchical models are expected to use less resources compared to baseline models.

\begin{table}[!h]
\tbl{Comparison of test accuracy and memory footprint between building block nets, HD-CNNs and hierarchical models on the ImageNet dataset. The testing mini-batch size is $50$.}
{\begin{tabular}{@{}cccc@{}} \toprule
\textbf{Model} & \textbf{Accuracy (\%)} & \textbf{Memory (MB)} & \textbf{Layers} \\\colrule
Baseline VGG-f & 55.32 & 546 & 7 \\ 
Hierarchical model & 50.07 & 458.66 & 7 \\ 
Adaptive transfer learning & 56.06 & 580.2 & 7 \\ 
Baseline VGG-16 layer & 67.7 & 4134 & 16 \\ 
HD-CNN + CE + PC & 68.66 & 6863 & 16 \\ \botrule
\end{tabular}}
\end{table}

Table 13 shows that memory requirements are approximately 14 $\times$ lower compared to the best model of the HD-CNN\cite{Yan2014HD-CNN:Recognition}. We also estimate that our models would require fewer computations while achieving improved performance when visual recognition task is scaled upwards to tens of thousands of object categories.

\section{Conclusion}

An automatic hierarchical clustering method is introduced which reduces parameters while simultaneously increasing classification accuracy.  This new approach borrows concepts from traditional divisive clustering techniques as well as confusion matrix dissimilarity linkage tree decomposition, to create an iterative method which methodically identifies cluster boundaries in a natural fashion. Hierarchical cluster boundary formation was tested on both MLP and CNN classifier frameworks, and shows significant benefit to the former, but marginal in the latter. It is hypothesized that other classification frameworks such as SVM and Bayes classifiers can also benefit from the hierarchical framework.  What is most intriguing is that the proposed strategy allows for virtually unlimited number of classes in any particular classification problem.

The proposed adaptive network selection framework, consisting of hierarchical models based on adaptive transfer learning, outperform single CNN models. The class assignment classifier network configuration is based on class confusion and composition statistics. As the complexity of classification problems increase, hierarchical models will offer significant benefits for large scale classification problems. Future work will demonstrate adaptive hierarchical clustering over multiple stages on the full ImageNet-$22K$ dataset. Use of ensembles to improve hierarchical classifier accuracy, data augmentations on imbalanced cluster to eradicate biases in hierarchical classifier predictions, and sharing initial layers among the class assignment classifiers will be used to improve hierarchical framework performance.

\section*{Acknowledgments}

We acknowledge NVIDIA for providing some of the GPU computing resources used in this research. We also thank Karthik Veerabhadra for his help in running some experiments.
\bibliography{main.bib}   % bibliography data in report.bib
\bibliographystyle{ws-ijprai.bst} 

\appendix

\section{Appendices}
Table 14, Table 15 \& Table 16 shows the estimation of parameter memory and data memory required by $CNN_{TL_{1}}$, $CNN_{TL_{3}}$ and $CNN_{TL_{5}}$ to make prediction on a test image. \\

\begin{table}[!h]
\tbl{Memory requirements for a Single CNN to make final prediction on a test image.}
{\begin{tabular}{@{}cm{1cm}cccm{1.5cm}ccccm{1.8cm}@{}}
\multicolumn{11}{c}{\textbf{ImageNet: Single CNN Classifier $CNN_{TL_{1}}$. MEMORY: 238.32 MB}} \\\toprule
\textbf{Layer} & \textbf{Function} & \multicolumn{3}{c}{\textbf{}} & \textbf{Data Memory (MB)} & \multicolumn{4}{c}{\textbf{}} & \textbf{Params Memory (MB)} \\ \colrule
0 & Input & 224 & 224 & 3 & 0.57 & 0 & 0 & 0 & 0 & 0 \\ 
1 & Conv1 & 54 & 54 & 64 & 0.71 & 11 & 11 & 3 & 64 & 0.09 \\  
1 & ReLU1 & 54 & 54 & 64 & 0.71 & 0 & 0 & 0 & 0 & 0 \\ 
1 & norm1 & 54 & 54 & 64 & 0.71 & 0 & 0 & 0 & 0 & 0 \\ 
1 & Pool1 & 27 & 27 & 64 & 0.18 & 0 & 0 & 0 & 0 & 0 \\  
2 & Conv2 & 27 & 27 & 256 & 0.71 & 5 & 5 & 64 & 256 & 1.56 \\ 
2 & ReLU2 & 27 & 27 & 256 & 0.71 & 0 & 0 & 0 & 0 & 0 \\ 
2 & norm2 & 27 & 27 & 256 & 0.71 & 0 & 0 & 0 & 0 & 0 \\ 
2 & Pool2 & 13 & 13 & 256 & 0.17 & 0 & 0 & 0 & 0 & 0 \\  
3 & Conv3 & 13 & 13 & 256 & 0.17 & 3 & 3 & 256 & 256 & 2.25 \\ 
3 & ReLU3 & 13 & 13 & 256 & 0.17 & 0 & 0 & 0 & 0 & 0 \\ 
4 & Conv4 & 13 & 13 & 256 & 0.17 & 3 & 3 & 256 & 256 & 2.25 \\ 
4 & ReLU4 & 13 & 13 & 256 & 0.17 & 0 & 0 & 0 & 0 & 0 \\  
5 & Conv5 & 13 & 13 & 256 & 0.17 & 3 & 3 & 256 & 256 & 2.25 \\ 
5 & ReLU5 & 13 & 13 & 256 & 0.17 & 0 & 0 & 0 & 0 & 0 \\ 
5 & Pool5 & 6 & 6 & 256 & 0.04 & 0 & 0 & 0 & 0 & 0 \\  
6 & Fc6 & 1 & 1 & 4096 & 0.02 & 6 & 6 & 256 & 4096 & 144 \\ 
6 & ReLU6 & 1 & 1 & 4096 & 0.02 & 0 & 0 & 0 & 0 & 0 \\  
7 & Fc7 & 1 & 1 & 4096 & 0.02 & 1 & 1 & 4096 & 4096 & 64 \\ 
7 & ReLu7 & 1 & 1 & 4096 & 0.02 & 0 & 0 & 0 & 0 & 0 \\  
8 & prob & 1 & 1 & 1000 & 0.01 & 1 & 1 & 4096 & 1000 & 15.63 \\  
\textbf{Total} & \multicolumn{4}{c}{\textbf{}} & \textbf{6.29} & \multicolumn{4}{c}{\textbf{}} & \textbf{232.03} \\ \botrule
\end{tabular}%
}
\end{table}
\begin{table}[!h]
\tbl{Memory requirements for a hierarchy classifier to make coarse category prediction on a test image}
{\begin{tabular}{@{}cm{1cm}cccm{1.5cm}ccccm{1.8cm}@{}}
\multicolumn{11}{c}{\textbf{ImageNet: hierarchy classifier $CNN_{TL_{3}}$. MEMORY: 103.36 MB}} \\ \toprule
\textbf{Layer} & \textbf{Function} & \multicolumn{3}{c}{} & \textbf{Data Memory (MB)} & \multicolumn{4}{c}{} & \textbf{Params Memory (MB)} \\ \colrule
0 & Input & 224 & 224 & 3 & 0.57 & 0 & 0 & 0 & 0 & 0 \\
1 & Conv1 & 54 & 54 & 64 & 0.71 & 11 & 11 & 3 & 64 & 0.09 \\ 
1 & ReLU1 & 54 & 54 & 64 & 0.71 & 0 & 0 & 0 & 0 & 0 \\ 
1 & norm1 & 54 & 54 & 64 & 0.71 & 0 & 0 & 0 & 0 & 0 \\ 
1 & Pool1 & 27 & 27 & 64 & 0.18 & 0 & 0 & 0 & 0 & 0 \\ 
2 & Conv2 & 27 & 27 & 256 & 0.71 & 5 & 5 & 64 & 256 & 1.56 \\ 
2 & ReLU2 & 27 & 27 & 256 & 0.71 & 0 & 0 & 0 & 0 & 0 \\
2 & norm2 & 27 & 27 & 256 & 0.71 & 0 & 0 & 0 & 0 & 0 \\ 
2 & Pool2 & 13 & 13 & 256 & 0.17 & 0 & 0 & 0 & 0 & 0 \\ 
3 & Conv3 & 13 & 13 & 256 & 0.17 & 3 & 3 & 256 & 256 & 2.25 \\ 
3 & ReLU3 & 13 & 13 & 256 & 0.17 & 0 & 0 & 0 & 0 & 0 \\
4 & Conv4 & 13 & 13 & 256 & 0.17 & 3 & 3 & 256 & 256 & 2.25 \\ 
4 & ReLU4 & 13 & 13 & 256 & 0.17 & 0 & 0 & 0 & 0 & 0 \\ 
5 & Conv5 & 13 & 13 & 256 & 0.17 & 3 & 3 & 256 & 256 & 2.25 \\ 
5 & ReLU5 & 13 & 13 & 256 & 0.17 & 0 & 0 & 0 & 0 & 0 \\
5 & Pool5 & 6 & 6 & 256 & 0.04 & 0 & 0 & 0 & 0 & 0 \\ 
6 & Fc6 & 1 & 1 & 2048 & 0.01 & 6 & 6 & 256 & 2048 & 72 \\ 
6 & ReLU6 & 1 & 1 & 2048 & 0.01 & 0 & 0 & 0 & 0 & 0 \\
7 & Fc7 & 1 & 1 & 2048 & 0.01 & 1 & 1 & 2048 & 2048 & 16 \\
7 & ReLu7 & 1 & 1 & 2048 & 0.01 & 0 & 0 & 0 & 0 & 0. \\
8 & prob & 1 & 1 & 89 & 0.01 & 1 & 1 & 2048 & 89 & 0.7 \\ 
\textbf{Total} & \multicolumn{4}{c}{} & \textbf{6.26} & \multicolumn{4}{c}{} & \textbf{97.1} \\ \botrule
\end{tabular}}
\end{table}

\begin{table}[!h]
\tbl{Memory requirements for a class assignment classifier to make fine category prediction on a test image.}
{\begin{tabular}{@{}cm{1cm}cccm{1.5cm}ccccm{1.8cm}@{}}
\multicolumn{11}{c}{\textbf{ImageNet: class assignment classifier $CNN_{TL_{5}}$. MEMORY: 54.84 MB (Worst Case )}} \\ \toprule
\textbf{Layer} & \textbf{Function} & \multicolumn{3}{c}{\textbf{}} & \textbf{Data Memory (MB)} & \multicolumn{4}{c}{\textbf{}} & \textbf{Params Memory (MB)} \\ \colrule
0 & Input & 224 & 224 & 3 & 0.57 & 0 & 0 & 0 & 0 & 0\\ 
1 & Conv1 & 54 & 54 & 64 & 0.71 & 11 & 11 & 3 & 64 & 0.09 \\  
1 & ReLU1 & 54 & 54 & 64 & 0.71 & 0 & 0 & 0 & 0 & 0 \\ 
1 & norm1 & 54 & 54 & 64 & 0.71 & 0 & 0 & 0 & 0 & 0 \\ 
1 & Pool1 & 27 & 27 & 64 & 0.18 & 0 & 0 & 0 & 0 & 0 \\
2 & Conv2 & 27 & 27 & 256 & 0.71 & 5 & 5 & 64 & 256 & 1.56 \\
2 & ReLU2 & 27 & 27 & 256 & 0.71 & 0 & 0 & 0 & 0 & 0 \\ 
2 & norm2 & 27 & 27 & 256 & 0.71 & 0 & 0 & 0 & 0 & 0 \\
2 & Pool2 & 13 & 13 & 256 & 0.17 & 0 & 0 & 0 & 0 & 0 \\ 
3 & Conv3 & 13 & 13 & 256 & 0.17 & 3 & 3 & 256 & 256 & 2.25 \\ 
3 & ReLU3 & 13 & 13 & 256 & 0.17 & 0 & 0 & 0 & 0 & 0 \\ 
4 & Conv4 & 13 & 13 & 256 & 0.17 & 3 & 3 & 256 & 256 & 2.25 \\
4 & ReLU4 & 13 & 13 & 256 & 0.17 & 0 & 0 & 0 & 0 & 0 \\ 
5 & Conv5 & 13 & 13 & 256 & 0.17 & 3 & 3 & 256 & 256 & 2.25 \\ 
5 & ReLU5 & 13 & 13 & 256 & 0.17 & 0 & 0 & 0 & 0 & 0 \\ 
5 & Pool5 & 6 & 6 & 256 & 0.04 & 0 & 0 & 0 & 0 & 0 \\ 
6 & Fc6 & 1 & 1 & 1024 & 0.01 & 6 & 6 & 256 & 1024 & 36 \\ 
6 & ReLU6 & 1 & 1 & 1024 & 0.01 & 0 & 0 & 0 & 0 & 0 \\ 
7 & Fc7 & 1 & 1 & 1024 & 0.01 & 1 & 1 & 1024 & 1024 & 4 \\ 
7 & ReLu7 & 1 & 1 & 1024 & 0.01 & 0 & 0 & 0 & 0 & 0 \\ 
8 & prob & 1 & 1 & 41 & 0.01 & 1 & 1 & 1024 & 41 & 0.16 \\ 
\textbf{Total} & \multicolumn{4}{c}{\textbf{}} & \textbf{6.26} & \multicolumn{4}{c}{\textbf{}} & \textbf{48.56} \\ \botrule
\end{tabular}}
\begin{tabnote}
Worst case indicates a network predicting classes in a cluster with maximum number of classes.\end{tabnote}
\end{table}

\end{document}